%% file: ex_article.tex
\documentclass[review,onefignum,onetabnum]{siamart220329}

\input{ex_shared}

\ifpdf
\hypersetup{
  pdftitle={nuGPR: GPU-Accelerated Gaussian Process Regression with Iterative Algorithms and Low-Rank Approximations},
  pdfauthor={Z. Zhao, and V. Sarin}
}
\fi

\begin{document}

\maketitle

% REQUIRED
\begin{abstract}
Gaussian Process Regression (GPR) is an important type of supervised machine learning model with inherent uncertainty measure in its predictions. We propose a new framework, nuGPR, to address the well-known challenge of high computation cost associated with GPR training. Our framework includes several ideas from numerical linear algebra to reduce the amount of computation in key steps of GPR, and we combine them to establish an end-to-end training algorithm. Specifically, we leverage the preconditioned conjugate gradient method to accelerate the convergence of the linear solves required in GPR. We exploit clustering in the input data to identify block-diagonal structure of the covariance matrix and subsequently construct low-rank approximations of the off-diagonal blocks. These enhancements significantly reduce the time and space complexity of our computations. In addition, unlike other frameworks that rely on exact differentiation, we employ numerical gradients to optimize the hyperparameters of our GPR model, further reducing the training cost by eliminating the need for backpropagation. Lastly, we leverage the CUDA Toolkit to efficiently parallelize the training procedure on NVIDIA GPUs. As a result, nuGPR reduces total training time by up to $2\times$ and peak memory consumption by up to $12\times$ on various synthetic and real-world datasets when compared to the best existing GPU-based GPR implementation.
\end{abstract}

% REQUIRED

% REQUIRED
\begin{MSCcodes}
% ref: https://mathscinet.ams.org/mathscinet/msc/msc2020.html
65Y05, % parallel implementation of numerical methods
60G15, % gaussian processes
65F10, % iterative numerical methods
65F55  % low-rank approximations
\end{MSCcodes}

\section{Introduction}
During recent years, Gaussian Process Regression (GPR) has gained a lot of attention in the machine learning field as a supervised probabilistic model. Essentially, it is a non-parametric model that describes a distribution over all possible functions that can fit a set of existing data \cite{wang2023intuitive}. Unlike most other machine learning algorithms, GPR provides quantification of the uncertainty associated with its predictions. This can be especially valuable in fields that require high confidence during the decision making process, such as risk management in finance and treatment management in healthcare.

The biggest limitation of GPR is its high computation cost and memory requirement associated with the computation and storage of the covariance matrix -- a square matrix denoting the correlation between any pair of input data. Such limitation is often amplified when large scale datasets are used; to address this, research on enhancing the scalability of GPR has witnessed considerable advancements in recent years. One category of methods to improve scalability leverages global or local approximation to the covariance matrix \cite{hensman2013gaussian, titsias2009variational, chalupka2013framework, gneiting2002compactly, quinonero2005unifying, wilson2015kernel, gramacy2008bayesian, yuksel2012twenty, rasmussen2001infinite, hinton2002training, deisenroth2015distributed, liu2020gaussian}. Global techniques typically approximate the covariance matrix by either using a subset of the training data to form a reduced matrix, or employing low-rank representations to create the Nystr\"{o}m approximation. Local methods, on the other hand, partition the input data into smaller subsets, solve each sub-problem, and combine the local solutions using a weighting scheme to make predictions. One issue these approximation techniques suffer from is the loss of information due to only a subset of data being used. Methods to average the output such as mixture of experts \cite{yuksel2012twenty} are used to enhance the accuracy of such approaches, but they come with their own complexity and issues, such as overfitting. The selection of the subset is also a concern, since a sub-optimal choice of data can easily pose bias to the model output.

Another category of methods to improve the scalability of GPR uses iterative methods \cite{saad2003iterative} instead of direct linear system solvers to optimize the hyperparameters \cite{cunningham2008fast, dong2017scalable, gardner2018product, pleiss2018constant, wilson2015kernel, gardner2018gpytorch}. Unlike traditional Gaussian Process inference, these approaches use iterative algorithms such as conjugate gradient (CG) or Lanczos tridiagonalization to compute approximations of the inverse and log determinant of the covariance matrix, which are required to compute the marginal log likelihood repeatedly during optimization. The dominant computation in these algorithms is matrix-vector multiplication -- a problem often perceived as more favorable for the computing architecture of GPUs than direct solvers such as Cholesky factorization \cite{volkov2008benchmarking, nath2011accelerating}.

In this paper, we introduce a framework, nuGPR, that significantly reduces the computational complexity and storage requirements for GPR training, while preserving its accuracy. Our approach relies on clustered input data, and is inspired by many aforementioned ideas such as iterative algorithms and low-rank representations. Furthermore, we include innovations to make our framework highly efficient and adaptable to cutting-edge GPU hardware. Below are our main contributions:

\begin{itemize}[left=0.25in]
    \item Based on clustered input data, our GPR training algorithm can limit expensive operations on only diagonal blocks of the covariance matrix, and approximate off-diagonal blocks by correlation among representative points of each cluster -- a strategy similar to other work leveraging inducing points \cite{chalupka2013framework, hensman2013gaussian, quinonero2005unifying, titsias2009variational, wilson2015kernel}. We demonstrate by synthetic and real-world datasets that our algorithm achieves comparable accuracy against exact computation, while offering considerable advantage in computation speed and memory consumption.

    \item We employ a preconditioned conjugate gradient (PCG) method for estimating both the inverse and log determinant of the covariance matrix \cite{saad1985parallel}. Specifically, to compute the log determinant, we use trace estimation \cite{avron2011randomized, hutchinson1989stochastic} combined with Pad\'e approximation \cite{pade1892representation}. The CG steps in both computation use the same preconditioner devised from Cholesky factorization on the diagonal blocks of the covariance matrix, similar to the idea introduced in \cite{concus1985block}. Our preconditioning technique accelerates the convergence of CG considerably, while incurring little overhead in practice.

    \item Our framework is designed with the goal of performing all matrix computation efficiently on GPUs. It is implemented in modern C++ and extensively utilizes NVIDIA's CUDA Toolkit and its various affiliated libraries. We ensure each dependency among the library routines is fine-tuned for optimal performance. As a result, nuGPR achieves up to $2\times$ reduction in computation speed and up to $12\times$ reduction in memory usage when compared to GPyTorch \cite{gardner2018gpytorch} -- the best existing implementation for GP inference tasks.
\end{itemize}

The paper is structured as follows. \Cref{background} introduces necessary background of GPR and reviews similar work in the domain. \Cref{methodology} establishes the theoretical foundation of our framework. \Cref{gpu} discusses our optimizations for efficient GPR training on the GPU. \Cref{experiment} presents the results that our framework achieves on various synthetic and real-world datasets. \Cref{discussion} elaborates on the experimental results and provides analysis of our framework. Finally, \Cref{conclusion} recaps the most important ideas we have covered in this paper.

\section{Background and Related Work} \label{background}
\subsection{Notations}
We use $\mathbf{X}_{\text{train}}$ to denote a training dataset of size $n_{\text{train}}$ in a $d$-dimensional space. Each training sample is denoted as $\mathbf{x}_i$, where $i$ means the $i$-th data point. Similarly, we will use $\mathbf{X}_{\text{test}}$ to represent a test set of size $n_{\text{test}}$. The ground truth labels for the training and testing set are denoted as $\mathbf{y}_{\text{train}}$ and $\mathbf{y}_{\text{test}}$, respectively.

The input $\mathbf{X}_{\text{train}}$ is partitioned into $n_c$ clusters $\{\mathbf{X}_1,\ldots, \mathbf{X}_{n_c}\}$. We employ a fixed cluster size of $b$ samples per cluster, hence the equality $n_{\text{train}} = bn_c$ always holds. We note that such assertions on the input data are an implementation choice rather than a limitation of our framework, as will be discussed in \Cref{gpu}.

\subsubsection*{Gaussian Process Regression}
Gaussian Process (GP) defines the target functions being modeled as a Gaussian distribution
\begin{equation}
    f(\mathbf{x}) \sim \mathcal{GP}(\mu(\mathbf{x}), k(\mathbf{x}, \mathbf{x}')),
\end{equation}
which is characterized by the mean function $\mu(\mathbf{x})$ and kernel function $k(\mathbf{x}, \mathbf{x}')$ for any input samples $\mathbf{x}$ and $\mathbf{x}'$. The kernel function is used to compute the variance between any pair of data points. We use $\mathbf{K}_{\mathbf{y}} = k(\mathbf{X}_{\text{train}}, \mathbf{X}_{\text{train}})$ to denote the covariance matrix, and $\hat{\mathbf{K}_{\mathbf{y}}} = \mathbf{K}_{\mathbf{y}} + \sigma^2 \mathbf{I}$ to denote the covariance matrix with added noise on the diagonal. $\mathbf{K}_* = k(\mathbf{X}_{\text{train}}, \mathbf{X}_{\text{test}})$ denotes the covariance matrix between the training and test samples, and $\mathbf{K}_{**} = k(\mathbf{X}_{\text{test}}, \mathbf{X}_{\text{test}})$ denotes the covariance matrix of the test samples.

There are a few common choices for the kernel function $k(\mathbf{x}, \mathbf{x}')$, and for consistency, we choose RBF as our kernel function for all GP computation. Specifically
\begin{equation}
    k(\mathbf{x}, \mathbf{x}') = \alpha \exp\left(-\frac{\|\mathbf{x} - \mathbf{x}'\|_2}{2\lambda^2}\right),
\end{equation}
where $\alpha$ denotes the output scale, $\lambda$ denotes the lengthscale, and $\|\mathbf{x} - \mathbf{x}'\|_2$ is the L2 norm between $\mathbf{x}$ and $\mathbf{x}'$. The 3-tuple $\theta = (\lambda, \sigma^2, \alpha)$ denotes all the hyperparameters of the GP model.

\subsubsection*{Train Forward Formulae}
Training a GP model involves optimizing its hyperparameters, such as the lengthscale and noise, by minimizing the negative marginal log likelihood \cite{williams2006gaussian}
\begin{equation} \label{loss function}
   \mathcal{L}(\theta| \mathbf{X}_{\text{train}}, \mathbf{y}_{\text{train}}) = \frac{1}{2}\mathbf{y}_{\text{train}}^\top \hat{\mathbf{K}_{\mathbf{y}}}^{-1} \mathbf{y}_{\text{train}} + \frac{1}{2}\log|\hat{\mathbf{K}_{\mathbf{y}}}| + \frac{n_{\text{train}}}{2}\log(2\pi),
\end{equation}
where the main computational steps are the linear solve of $\hat{\mathbf{K}_{\mathbf{y}}}$ for the right-hand-side $\mathbf{y}_{\text{train}}$ and the log determinant of $\hat{\mathbf{K}_{\mathbf{y}}}$. The last term depends on only $n_{\text{train}}$ and is trivial to compute.

A common approach for the above optimization procedure is to use the Adam optimizer \cite{kingma2014adam}, which takes as input the gradients of $\mathcal{L}$ w.r.t. each individual hyperparameter -- namely $\frac{\partial\mathcal{L}}{\partial \lambda}$, $\frac{\partial\mathcal{L}}{\partial \sigma^2}$ and $\frac{\partial\mathcal{L}}{\partial \alpha}$, and updates the set of hyperparameters $\theta' = (\lambda', \sigma'^2, \alpha')$ based on a pre-defined learning rate $\gamma$. While the analytical forms of these gradients are available, we skip presenting the formulae for evaluating them, since our usage of numerical gradients described in \Cref{numerical algorithm} helps us avoid computing the gradients directly.

\subsubsection*{Posterior Forward Formulae}
The predictive posterior distribution for the test set $\mathbf{X}_{\text{test}}$, denoted as $p(f(\mathbf{X}_{\text{test}}) | \mathbf{X}_{\text{train}}, \mathbf{y}_{\text{train}})$, can be expressed as the predictive mean vector and covariance matrix
\begin{align}
    \mu_{\text{pred}} &= \mathbf{K}_*^\top \hat{\mathbf{K}_{\mathbf{y}}}^{-1}\mathbf{y}_{\text{train}}, \\
    \mathbf{K}_{\text{pred}} &= \mathbf{K}_{**} - \mathbf{K}_*^{\top} \hat{\mathbf{K}_{\mathbf{y}}}^{-1} \mathbf{K}_*,
\end{align}
which are fundamental to the evaluation of the GP inference results. Here, a zero prior mean is assumed for $\mu_{\text{pred}}$. The Root Mean Squared Error (RMSE) score
\begin{equation}
    \sqrt{\|\mu_{\text{pred}} - \mathbf{y}_{\text{test}}\|_2}
\end{equation}
can be used to characterize the accuracy of $\mu_{\text{pred}}$, and the confidence region given by
\begin{equation}
    (\mu_{\text{pred}} - 2\mathbf{s}_{\text{pred}},\ \mu_{\text{pred}} + 2\mathbf{s}_{\text{pred}}),
\end{equation}
is an important measure about the uncertainty of the GP model on its predictions. Here, $\mathbf{s}_{\text{pred}}$ is a vector obtained by taking the square root of each diagonal element of $\mathbf{K}_{\text{pred}}$.

\subsection{Challenges in GPR Training}
A major challenge of using Gaussian Process models is the high time and space complexity associated with the linear solve and log determinant computation, as highlighted in \Cref{loss function}. In particular, for a training set of size $n_{\text{train}}$, the computation steps incur $\mathcal{O}(n_{\text{train}}^3)$ time and $\mathcal{O}(n_{\text{train}}^2)$ space complexity, which can quickly become impractical as the size of the dataset grows. Therefore, reducing the amount of computation during the training process, as well as finding a more memory efficient way to store the covariance matrix, has become increasingly important in GPR training.

For certain data distributions, an additional challenge is that the covariance matrix can be near-singular or ill-conditioned. This may be caused by artifacts in the dataset, where multiple samples are too close to each other, yielding nearly identical rows in the covariance matrix. These artifacts can create numerical instabilities for direct linear solvers and iterative solvers, degrading the computational results. The model can become more accurate if such instabilities can be avoided from the data, operations with good numerical stability are used, or methods such as preconditioning are introduced to deal with these issues.

\subsection{Related Work} \label{related work}
\subsubsection*{Approximation Techniques}
The simplest approximation that can be used to reduce computation is Subset-of-Data (SoD), where a subset of the training data is selected for GP inference. It can achieve a complexity of $\mathcal{O}(n_{\text{sub}}^3)$, where $n_{\text{sub}}$ is the size of the subset. The selection of the subset can usually be one of three methods \cite{liu2020gaussian}. The first is random selection. The second is selecting centroids after clustering the data into a number of subsets. The third is employing active learning criteria \cite{lawrence2002fast, seeger2003bayesian, keerthi2005matching}. More advanced techniques that aim to reduce the computation complexity of GP inference, such as SoR \cite{smola2000sparse}, FITC \cite{snelson2005sparse}, SVGP \cite{hoffman2013stochastic, hoang2015unifying}, use the idea of inducing points. Selection techniques for the inducing points \cite{keerthi2005matching, smola2000sparse, seeger2003fast, snelson2005sparse} are key to the viability of these methods.

\subsubsection*{Sparse Kernel Methods}
Another category of techniques to speed up GP computations construct a sparse representation of the covariance matrix, and they are commonly referred to as sparse kernels \cite{melkumyan2009sparse}. Here, the covariance matrix is obtained through a special kernel which can result in more zero entries and thus reduce the computation workload during inference. One key challenge is to ensure the matrix is still positive semi-definite after pruning those entries \cite{buhmann2001new, wendland2004scattered}.

\subsubsection*{Iterative Algorithms in GPR Training}
Traditional GPR training uses direct methods such as Cholesky factorization for the linear solve and log determinant terms shown in \Cref{loss function}. The factorization routine also has $\mathcal{O}(n_{\text{train}}^3)$ time complexity, which makes it difficult to scale GP models, especially if the exact $n_{\text{train}}\times n_{\text{train}}$ covariance matrix is needed explicitly. An alternative approach is to use iterative algorithms such as conjugate gradient, which primarily involve matrix-vector multiplications \cite{shen2005fast, morariu2008automatic, chalupka2013framework, saatcci2012scalable, cunningham2008fast, wilson2015kernel}. Although these algorithms have a theoretical complexity of $\mathcal{O}(n_{\text{train}}^3)$ as well, in practice they can often take $k \ll n_{\text{train}}$ iterations to converge, which lowers the time complexity of the linear solve to $\mathcal{O}(kn_{\text{train}}^2)$. Moreover, preconditioned conjugate gradient (PCG) is often used in practice instead of CG to accelerate the rate of convergence and to tackle ill-conditioned covariance matrices successfully \cite{cutajar2016preconditioning}.

\subsubsection*{GPU Acceleration for GPR}
Most existing efforts for GPR have focused on CPU-based computations. Recently, there has been a growing recognition of the advantages of leveraging GPUs for GP inference, thanks to the architectural benefits they can provide. This shift has led to the development of GPU-accelerated GPR frameworks. Notably, GPyTorch \cite{gardner2018gpytorch} and GPFlow \cite{GPflow2017} are two well-known projects that harness the power of GPUs. They utilize intermediate software libraries with CUDA support -- PyTorch \cite{paszke2019pytorch} for GPyTorch and TensorFlow \cite{tensorflow2015-whitepaper} for GPFlow -- to significantly accelerate GP inference on NVIDIA GPUs.

An alternative to these packages is to write a GP framework that depends natively on NVIDIA's CUDA Toolkit \cite{luebke2008cuda}. One notable overhead of frameworks like PyTorch and TensorFlow is their reliance on intermediate layers, such as JIT (just-in-time) compilation, to translate Python code to CUDA APIs. In contrast, the CUDA Toolkit is built directly on top of C++, and provides native access to computation kernels on the GPU. It also includes state-of-the-art linear algebra libraries, such as cuBLAS\footnote{Software available on NVIDIA website: \url{https://docs.nvidia.com/cuda/cublas/index.html}} and cuSOLVER\footnote{Software available on NVIDIA website: \url{https://docs.nvidia.com/cuda/cusolver/index.html}}, that provide advanced routines for matrix operations and numerical linear algebra procedures like Cholesky factorization. Despite the potential benefits, this approach to directly use the CUDA Toolkit for GP has not seen much advancement in recent years.

\section{Methodology} \label{methodology}
We present nuGPR, a GPU-based Gaussian Process Regression framework based on both (nu)merical linear algebra and the (nu)merical gradient optimization technique. The following section details the main theoretical considerations of our framework.

\subsection{Numerical Algorithms} \label{numerical algorithm}
\subsubsection*{CG in the Computation of $\hat{\mathbf{K}_{\mathbf{y}}}^{-1}\mathbf{y}_{\text{train}}$}
Similar to many previous works, we employ CG when computing the $\mathbf{y}_{\text{train}}^\top \hat{\mathbf{K}_{\mathbf{y}}}^{-1}\mathbf{y}_{\text{train}}$ term in \Cref{loss function}. More specifically, to compute $\hat{\mathbf{K}_{\mathbf{y}}}^{-1}\mathbf{y}_{\text{train}}$, we are essentially solving $\hat{\mathbf{K}_{\mathbf{y}}}\mathbf{u} = \mathbf{y}_{\text{train}}$ for $\mathbf{u}$. Because $\hat{\mathbf{K}_{\mathbf{y}}}$ is symmetric positive-definite (SPD), CG becomes a great candidate for solving the system. In our experiments, the number of iterations $k$ needed for CG to converge usually satisfies $2\leq k\leq 10$ even for large datasets we have considered, thanks to the use of preconditioning as discussed in \Cref{preconditioning}. After computing $\hat{\mathbf{K}_{\mathbf{y}}}^{-1}\mathbf{y}_{\text{train}}$, only a simple dot product with $\mathbf{y}_{\text{train}}$ is required to obtain the result of this term. Such process alone significantly reduces the computation complexity from $\mathcal{O}(n_{\text{train}}^3)$ to $\mathcal{O}(kn_{\text{train}}^2)$. After our low-rank approximation described in \Cref{low rank}, more reduction in this complexity can be achieved. In addition, we observe the same effect discussed in \cite{gardner2018gpytorch}, where PCG improves the numeric stability of the solve compared to Cholesky factorization of the matrix.

\subsubsection*{Computation of $\log|\hat{\mathbf{K}_{\mathbf{y}}}|$}
For the computation of the log determinant term $\log|\hat{\mathbf{K}_{\mathbf{y}}}|$ in \Cref{loss function}, we take advantage of several properties of the covariance matrix $\hat{\mathbf{K}_{\mathbf{y}}}$, which is inherently SPD. First, we can transform $\log|\hat{\mathbf{K}_{\mathbf{y}}}|$ into
\begin{equation} \label{log to trace}
    \log|\hat{\mathbf{K}_{\mathbf{y}}}| = \text{tr}(\log(\hat{\mathbf{K}_{\mathbf{y}}})),
\end{equation}
and then approximate the right-hand-side using trace estimation \cite{avron2011randomized, hutchinson1989stochastic}
\begin{equation} \label{trace estimation}
    \text{tr}(\log(\hat{\mathbf{K}_{\mathbf{y}}})) \approx \frac{1}{m} \sum_i^m \mathbf{z}_i^\top \log(\hat{\mathbf{K}_{\mathbf{y}}}) \mathbf{z}_i,
\end{equation}
where $m$ is the number of column vectors in the estimation, and $\mathbf{z}_i$ are Hutchinson vectors \cite{hutchinson1989stochastic} whose entries are randomly chosen as either $1$ or $-1$. Furthermore, for the computation of $\log(\hat{\mathbf{K}_{\mathbf{y}}})$, we use a polynomial expansion of the matrix $\hat{\mathbf{K}_{\mathbf{y}}}$, similar to the idea mentioned in \cite{zhang2007approximate}. However, instead of using Taylor series with a maximum degree of $30$, we employ a 2-2 Pad\'e approximation for matrix logarithms
\begin{equation} \label{PQ}
    \log(\hat{\mathbf{K}_{\mathbf{y}}}) \approx P(\hat{\mathbf{K}_{\mathbf{y}}})Q^{-1}(\hat{\mathbf{K}_{\mathbf{y}}}) = (3 \hat{\mathbf{K}_{\mathbf{y}}}^2 - 3 \mathbf{I}) (\hat{\mathbf{K}_{\mathbf{y}}}^2 + 4 \hat{\mathbf{K}_{\mathbf{y}}} + \mathbf{I})^{-1} .
\end{equation}

Note that $Q(\hat{\mathbf{K}_{\mathbf{y}}})$ preserves the SPD property of the original $\hat{\mathbf{K}_{\mathbf{y}}}$. Therefore, combining the above, in order to compute $\text{tr}(\log(\hat{\mathbf{K}_{\mathbf{y}}}))$, we use CG to first solve for $Q(\hat{\mathbf{K}_{\mathbf{y}}})^{-1}\mathbf{z}_i$ for each $i$, multiply the result with $P(\hat{\mathbf{K}_{\mathbf{y}}})$, and finally compute the dot product against vector $\mathbf{z}_i$. Implementation wise, since there are $m$ columns on the right-hand-side, we adopt a batch CG algorithm similar to the idea in \cite{gardner2018gpytorch} to compute all $m$ solves simultaneously, better utilizing the parallel capability of the GPU.

The above approach reduces the complexity for computing the log determinant term from $\mathcal{O}(n_{\text{train}}^3)$ to $\mathcal{O}(kmn_{\text{train}}^2)$ for some iteration number $k$. As described in \Cref{preconditioning}, this approach can easily be extended to the case when $\hat{\mathbf{K}_{\mathbf{y}}}$ is preconditioned with the same block-diagonal matrix we use as a preconditioner when solving $\hat{\mathbf{K}_{\mathbf{y}}}\mathbf{u} = \mathbf{y}_{\text{train}}$ via PCG, which results in $k$ being very small. Similar to the $\hat{\mathbf{K}_{\mathbf{y}}}^{-1}\mathbf{y}_{\text{train}}$ case, our low-rank approximation in \Cref{low rank} further reduces this time complexity.

\subsubsection*{Numerical Gradient}
To avoid explicitly computing the gradients of $\mathcal{L}(\theta | \mathbf{X}, \mathbf{y})$ for each hyperparameter with automatic differentiation, we use numerical gradient to approximate these quantities instead. Specifically, given a step size $\Delta \theta_i$ in the $i$-th hyperparameter, we compute the gradient
\begin{equation}\label{numerical gradient}
    \frac{\partial \mathcal{L}(\theta | \mathbf{X}_{\text{train}}, \mathbf{y}_{\text{train}})}{\partial \theta_i} = \frac{\mathcal{L}(\theta' | \mathbf{X}_{\text{train}}, \mathbf{y}_{\text{train}}) - \mathcal{L}(\theta | \mathbf{X}_{\text{train}}, \mathbf{y}_{\text{train}})}{\Delta\theta_i},
\end{equation}
where $\theta'$ is obtained by substituting $\theta_i$ with $\theta_i + \Delta\theta_i$. To ensure the obtained gradient value is as close to the real gradient value as possible, we continuously reduce $\Delta\theta_i$ by half until the difference between two consecutive gradients is smaller than a given threshold, usually between $10^{-6}$ and $10^{-2}$. We note that above is commonly used approach in similar numerical optimization problems \cite{nocedal1999numerical, lecun1998gradient}.

\subsection{Preconditioning} \label{preconditioning}
To further reduce the computation cost associated with CG, we introduce a preconditioning strategy that can reduce the number of iterations $k$ for CG to converge. The strategy is based on the assumption that the block-diagonal matrix $\mathbf{K}_{\text{diag}}$, comprised of all $b\times b$ blocks on the diagonal of $\hat{\mathbf{K}_{\mathbf{y}}}$, reflects all the significant entries in the covariance matrix. In other words, we expect any large value of $k(\mathbf{x},\mathbf{x}')$ to be contributed by $\mathbf{x}$ and $\mathbf{x}'$ of the same cluster. 

Our preconditioning algorithm is based on the block Jacobi preconditioner \cite{cutajar2016preconditioning}; however, instead of directly approximating the matrix inverse, we perform Cholesky factorization on each diagonal block $(\mathbf{K}_{\text{diag}})_i$ of the covariance matrix, and assemble the resulting matrix blocks as
\begin{align}
    \mathbf{R} & = \text{diag}(\mathbf{R}_i), \\
    \mathbf{R}_i^\top \mathbf{R}_i & = (\mathbf{K}_{\text{diag}})_i,\ i = 1, \dots, n_c, \label{cholesky factorization}
\end{align}
with the intention of using $\mathbf{R}^{-1}$ to precondition the original matrix $\hat{\mathbf{K}_{\mathbf{y}}}$. Our motivation is that if $\mathbf{K}_{\text{diag}}$ approximates the entries of $\hat{\mathbf{K}_{\mathbf{y}}}$ well enough, $\mathbf{R}^{-\top}\hat{\mathbf{K}_{\mathbf{y}}}\mathbf{R}^{-1}$ will be close to the identity, which improves the condition of the PCG solve and allows it to converge rapidly \cite{saad1985parallel}.

For the solves $Q(\hat{\mathbf{K}_{\mathbf{y}}})^{-1}\mathbf{z}_i$ required by \Cref{PQ}, it would be desirable to reuse the above $\mathbf{R}$ to speed up the convergence, since the condition number of $Q(\hat{\mathbf{K}_{\mathbf{y}}})$ is increased due to the quadratic matrix polynomial, making an effective preconditioning strategy more important. Therefore, we describe the process of computing $\log|\hat{\mathbf{K}_{\mathbf{y}}}|$ with $\mathbf{R}^{-\top}\hat{\mathbf{K}_{\mathbf{y}}}\mathbf{R}^{-1}$ as the preconditioned system to solve against, and compensating the final result for the log determinant. Since $\mathbf{R}= \text{diag}(\mathbf{R}_i)$, we have
\begin{equation}
    |\mathbf{R}| = \prod_i^{n_c} |\mathbf{R}_i|,
\end{equation}
and therefore
\begin{equation} \label{RKR}
    |\mathbf{R}^{-\top} \hat{\mathbf{K}_{\mathbf{y}}} \mathbf{R}^{-1}| = (\prod_i^{n_c} |\mathbf{R}_i|^2)^{-1} |\hat{\mathbf{K}_{\mathbf{y}}}|,
\end{equation}
which, after combining with \Cref{log to trace} and \cref{trace estimation}, gives us
\begin{equation} \label{logdet}
    \log|\hat{\mathbf{K}_{\mathbf{y}}}| = 2\sum_i^{n_c} \log |\mathbf{R}_i| + \frac{1}{m}\sum_i^m \mathbf{z}_i^\top \log(\mathbf{R}^{-\top}\hat{\mathbf{K}_{\mathbf{y}}}\mathbf{R}^{-1}) \mathbf{z}_i.
\end{equation}

Essentially, we transform the original system of $Q(\hat{\mathbf{K}_{\mathbf{y}}})$ into $Q(\mathbf{R}^{-\top}\hat{\mathbf{K}_{\mathbf{y}}}\mathbf{R}^{-1})$, which is much better-conditioned. This allows us to keep the number of iteration $k$ small when computing the Pad\'e approximation, and reduce the overall computation cost incurred by PCG.

In practice, the computation of $\mathbf{R}$ requires $\mathcal{O}(b^3n_c)$ time to complete a total of $n_c$ Cholesky factorizations. This is much less than the $\mathcal{O}(n_{\text{train}}^3)$ factorization of $\hat{\mathbf{K}_{\mathbf{y}}}$, but is still far from trivial. We alleviate this by computing $\mathbf{R}$ only once for the entire training epoch. Specifically, in addition to using $\mathbf{R}$ for all PCG solves, we note that $\mathbf{R}$ is also suitable for computing the numerical gradient in \Cref{numerical gradient}. We expect the covariance matrix evaluated at $\theta_i + \Delta\theta_i$ for the $i$-th hyperparameter to closely resemble $\hat{\mathbf{K}_{\mathbf{y}}}$, making $\mathbf{R}^{-1}$ still a viable preconditioner to retain the computational efficiency. 

\subsection{Low-Rank Approximation for Off-Diagonal Blocks} \label{low rank}
Recall that the partitioning $\mathbf{X}_{\text{train}} = \{\mathbf{X}_1,\ldots, \mathbf{X}_{n_c}\}$ splits the training set into clusters of $b$ samples each. Now, denote the covariance matrix between clusters $\mathbf{X}_i$ and $\mathbf{X}_j$ as $\mathbf{K}_{ij} = k(\mathbf{X}_i, \mathbf{X}_j)$. We can view the original $\hat{\mathbf{K}_{\mathbf{y}}}$ as an $n_c\times n_c$ grid consisting of $b\times b$ sized blocks, where the diagonal blocks form $\mathbf{K}_{\text{diag}}$, and the off-diagonal blocks are $\mathbf{K}_{ij}$ for all $i\neq j$.

Continuing the idea that $\mathbf{K}_{\text{diag}}$ contains the most significant entries of $\hat{\mathbf{K}_{\mathbf{y}}}$, we argue that lowering the precision of any off-diagonal $\mathbf{K}_{ij}$ -- for example, using a low-rank approximation $\mathbf{K}_{ij}'$ in place of $\mathbf{K}_{ij}$ -- does not significantly impact the accuracy of GPR training. Specifically, we leverage the following rank-$1$ approximation in our implementation
\begin{equation} \label{odblock}
    \mathbf{K}_{ij}' = k(\mathbf{r}_i, \mathbf{r}_j)\cdot \mathbf{1} \cdot \mathbf{1}^\top,
\end{equation}
where $\mathbf{r}_i$ and $\mathbf{r}_j$ denote the \textit{representative points} for cluster $i$ and $j$, respectively. The intuition is that for any $\mathbf{x}_i\in \mathbf{X}_i$ and $\mathbf{x}_j\in \mathbf{X}_j$, we approximate the kernel value $k(\mathbf{x}_i, \mathbf{x}_j)\approx k(\mathbf{r}_i, \mathbf{r}_j)$, since $\mathbf{r}_i$ and $\mathbf{r}_j$ are \textit{representative} of their own clusters. As a result, all entries in $\mathbf{K}_{ij}$ are replaced by the same scalar $k(\mathbf{r}_i, \mathbf{r}_j)$, giving us the approximation $\mathbf{K}_{ij}'$ as above. We use $\hat{\mathbf{K}_{\mathbf{y}}}'$ to denote the approximated matrix with all off-diagonal blocks $\mathbf{K}_{ij}$ of $\hat{\mathbf{K}_{\mathbf{y}}}$ replaced by $\mathbf{K}_{ij}'$.

There are several propositions we hope to demonstrate through our experiments in \Cref{experiment}. First, the methods for selecting the above representative points can vary depending on the context of the dataset, but all of them perform well to embed cluster information in the training dataset. Second, while one can choose any $1\leq \text{rank}(\mathbf{K}_{ij}') < b$ in theory, setting $\text{rank}(\mathbf{K}_{ij}') = 1$ uniformly already suffices to preserve the model's prediction accuracy.

\subsubsection*{Performance Improvement}
We describe how the low-rank approximation in \Cref{odblock} improves both time and space complexity for the training process of nuGPR.

One of the dominant factors of the training time is the PCG step to compute the Pad\'e approximation in \Cref{PQ}. Each iteration of PCG requires the left-multiplication of the linear system $\mathbf{R}^{-\top}\hat{\mathbf{K}_{\mathbf{y}}}\mathbf{R}^{-1}$ to the search direction $\mathbf{D}$, which is $n_{\text{train}}\times m$ considering that we are batching $m$ solves for vectors $\mathbf{z}_1,\ldots, \mathbf{z}_m$. Without the low-rank approximation, this multiplication takes $\mathcal{O}(mn_{\text{train}}^2)$ time, with the bottleneck being the multiplication of $\hat{\mathbf{K}_{\mathbf{y}}}$ to the intermediate result $\mathbf{R}^{-1} \mathbf{D}$.

With our approximation, let $\hat{\mathbf{K}_{\mathbf{y}}}' = \mathbf{K}_{\text{diag}} + \mathbf{K}_{\text{offdiag}}$, with $\mathbf{K}_{\text{offdiag}}$ consisting of the off-diagonal blocks $\mathbf{K}_{ij}'$ defined in \Cref{odblock}. We have
\begin{equation} \label{odsplit}
    \mathbf{R}^{-\top} \hat{\mathbf{K}_{\mathbf{y}}}' \mathbf{R}^{-1} = \mathbf{R}^{-\top} \mathbf{K}_{\text{diag}} \mathbf{R}^{-1} + \mathbf{R}^{-\top} \mathbf{K}_{\text{offdiag}} \mathbf{R}^{-1},
\end{equation}
where the diagonal blocks of $\mathbf{R}^{-\top} \mathbf{K}_{\text{diag}} \mathbf{R}^{-1}$ are simply
\begin{equation}
    \mathbf{R}_i^{-\top} (\mathbf{K}_{\text{diag}})_i \mathbf{R}_i^{-1},
\end{equation}
and the off-diagonal blocks of $\mathbf{R}^{-\top} \mathbf{K}_{\text{offdiag}} \mathbf{R}^{-1}$ can be expressed in the following form
\begin{align}
    & k(\mathbf{r}_i, \mathbf{r}_j)\cdot \mathbf{u}_i\cdot \mathbf{u}_j^\top, \\
    & \mathbf{u}_i = \mathbf{R}_i^{-\top}\cdot \mathbf{1},\ i = 1,\ldots, n_c.
\end{align}

All vectors $\mathbf{u}_i$ can be pre-computed, which requires $\mathcal{O}(b^2n_c)$ work in total. Once $\mathbf{u}_i$'s are available, applying $\mathbf{R}^{-\top} \hat{\mathbf{K}_{\mathbf{y}}}' \mathbf{R}^{-1}$ to each column of $\mathbf{D}$ requires $\mathcal{O}(b^2n_c)$ work for the block-diagonal part, and $\mathcal{O}(bn_c^2)$ for the off-diagonal part. This leads to a total time complexity of $\mathcal{O}(mb^2n_c + mbn_c^2)$ to complete the multiplication for the entire matrix $\mathbf{D}$. And since $b\gg n_c$ in all our experiments, the complexity can be simplified to just $\mathcal{O}(mb^2n_c)$.

Finally, in contrast to traditional GPR frameworks who often store the complete covariance matrix using $\mathcal{O}(n_{\text{train}}^2)$ storage, we only need to store the pair of matrices
\begin{equation}
    (\mathbf{K}_{\text{diag}}, \mathbf{K}_{\text{rep}})
\end{equation}
to encode the matrix $\hat{\mathbf{K}_{\mathbf{y}}}'$. Here, $\mathbf{K}_{\text{rep}}$ is the covariance matrix between all representative points $\mathbf{r}_1, \ldots, \mathbf{r}_{n_c}$. This storage format requires only $\mathcal{O}(b^2n_c + n_c^2)$ memory, which simplifies as $\mathcal{O}(b^2n_c)$ per our above discussion. So, to summarize, using the low-rank $\hat{\mathbf{K}_{\mathbf{y}}}'$ as the approximated covariance matrix significantly improves both the per-iteration time and space complexity of PCG.

\subsubsection*{More Optimization for Numerical Gradient}
For selected steps in numerical gradient, we can further simplify the computation of $\mathbf{R}^{-\top} \mathbf{K}_{\text{diag}} \mathbf{R}^{-1}$ in the above discussion. Suppose the model currently has hyperparameters $\theta_0 = (\lambda_0, \sigma_0^2, \alpha_0)$, and let $(\hat{\mathbf{K}_{\mathbf{y}}})_0$ denote the covariance matrix obtained from $\theta_0$. We compute \Cref{loss function} under $\theta_0$ as a baseline for varying each hyperparameter, and we obtain $\mathbf{R}^{-1}$ by factorizing $(\mathbf{K}_{\text{diag}})_0$. Now
 \begin{equation} \label{linear1}
    \mathbf{R}^{-\top} (\mathbf{K}_{\text{diag}})_0 \mathbf{R}^{-1} \mathbf{D} = \mathbf{D},
\end{equation}
thus the first term that originally dominates the complexity in \Cref{odsplit} becomes trivial to compute.

Now, suppose we vary the noise parameter by step size $\Delta\sigma^2$ and obtain $\theta_{\sigma_0^2 + \Delta\sigma^2} = (\lambda_0, \sigma_0^2 + \Delta\sigma^2, \alpha_0)$. It is straightforward to derive that $(\mathbf{K}_{\text{diag}})_{\sigma_0^2 + \Delta\sigma^2} = (\mathbf{K}_{\text{diag}})_0 + \Delta\sigma^2 \mathbf{I}$, and thus
\begin{equation} \label{linear2}
    \mathbf{R}^{-\top} (\mathbf{K}_{\text{diag}})_{\sigma_0^2 + \Delta\sigma^2} \mathbf{R}^{-1} \mathbf{D} = \mathbf{D} + \Delta\sigma^2 \mathbf{R}^{-\top} \mathbf{R}^{-1} \mathbf{D},
\end{equation}
which reduces the required number of block-diagonal left matrix multiplications. Specifically, on the left-hand-side we need to left-multiply $\mathbf{D}$ by $\mathbf{R}^{-1}$ first, then $(\mathbf{K}_{\text{diag}})_{\sigma_0^2 + \Delta\sigma^2}$, and finally $\mathbf{R}^{-\top}$, but on the right-hand-side, we only need to left-multiply by $\mathbf{R}^{-1}$ followed by $\mathbf{R}^{-\top}$.

A similar calculation can be performed when we vary the output scale parameter by $\Delta\alpha$: $\theta_{\alpha_0 + \Delta\alpha} = (\lambda_0, \sigma_0^2, \alpha_0 + \Delta\alpha)$. In such case, computing the ratio $r_\alpha = \frac{\Delta\alpha}{\alpha}$ allows us to write $(\mathbf{K}_{\text{diag}})_{\alpha_0 + \Delta\alpha} = (1 + r_{\alpha})(\mathbf{K}_{\text{diag}})_0 - \sigma^2 r_\alpha \mathbf{I}$, and a similar reduction in left matrix multiplications can be achieved
\begin{equation} \label{linear3}
    \mathbf{R}^{-\top} (\mathbf{K}_{\text{diag}})_{\alpha_0 + \Delta\alpha} \mathbf{R}^{-1} \mathbf{D} = (1 + r_\alpha) \mathbf{D} - \sigma^2 r_\alpha \mathbf{R}^{-\top} \mathbf{R}^{-1} \mathbf{D}.
\end{equation}

The takeaway is that except for varying the lengthscale $\lambda$, all steps in numerical gradient can be effectively optimized by reducing the number of instances of computing $\mathbf{R}^{-\top} \mathbf{K}_{\text{diag}} \mathbf{R}^{-1} \mathbf{D}$ to the minimum possible needed. Note, this optimization is applied to the dominant factor in each iteration of PCG, so essentially, we have reduced the constant factor in the $\mathcal{O}(mb^2n_c)$ time complexity in our above analysis.

\subsubsection*{Positive-Definiteness of $\hat{\mathbf{K}_{\mathbf{y}}}'$}
In order for CG to be applicable to $\hat{\mathbf{K}_{\mathbf{y}}}'$, we need to ensure the matrix is SPD. While such property is not theoretically guaranteed by our choice of approximation in \Cref{odblock}, we note that the issue can be addressed if we consider the Cholesky factorization of $\hat{\mathbf{K}_{\mathbf{y}}}'$, since the success of this factorization is a sufficient and necessary condition for the matrix to be SPD. For the scenarios when the factorization of $\hat{\mathbf{K}_{\mathbf{y}}}'$ fails, one can add small values to the diagonal terms of $\hat{\mathbf{K}_{\mathbf{y}}}'$ on the fly during the factorization process to make sure the matrix becomes SPD \cite{williams2006gaussian}.

Another approach to construct $\hat{\mathbf{K}_{\mathbf{y}}}'$ to be SPD, considering its block-diagonal structure, is to add compensation to its diagonal blocks. Recall the $n_c\times n_c$ covariance matrix $\mathbf{K}_{\text{rep}}$, and let $\lambda_0 > 0$ be its smallest eigenvalue. The following matrix
\begin{equation} \label{krep}
    \mathbf{K}_{\text{rep}} - \lambda_0 \mathbf{I}_{n_c}
\end{equation}
has a smallest eigenvalue of $0$, thus it is positive semi-definite. Now, for the matrix $\mathbf{K}_{\text{offdiag}}$ in \Cref{odsplit}, all of its diagonal blocks are filled with $0$ by definition. We add the following \textit{compensation value}
\begin{equation}
    k(\mathbf{r}_i, \mathbf{r}_i) - \lambda_0
\end{equation}
to each of its $i$-th diagonal block. The resulting matrix, denoted as $\mathbf{K}_{\text{offdiag}}'$, is positive semi-definite, since its eigenvalues are either $0$ or the same as the eigenvalues of the matrix in \Cref{krep}. Therefore, the \textit{compensated covariance matrix}
\begin{equation}\label{kypp}
    \hat{\mathbf{K}_{\mathbf{y}}}'' = \mathbf{K}_{\text{diag}} + \mathbf{K}_{\text{offdiag}}'
\end{equation}
is SPD, since it is the sum of an SPD matrix and a symmetric positive semi-definite matrix. We conclude that $\hat{\mathbf{K}_{\mathbf{y}}}''$ is a good candidate to apply CG solve in the training process.

\subsubsection*{Effectiveness of Preconditioning}
Finally, we note that \Cref{kypp} also gives us insights about how our preconditioner $\mathbf{R}^{-1}$ improves the convergence of CG. Consider the preconditioned matrix
\begin{equation}\label{precond kypp}
    \mathbf{R}^{-\top} \hat{\mathbf{K}_{\mathbf{y}}}'' \mathbf{R}^{-1} = \mathbf{I} + \mathbf{R}^{-\top} \mathbf{K}_{\text{offdiag}}' \mathbf{R}^{-1},
\end{equation}
because $\mathbf{K}_{\text{offdiag}}'$ is at most rank $n_c$, the same bound applies to $\mathbf{R}^{-\top} \mathbf{K}_{\text{offdiag}}' \mathbf{R}^{-1}$. This means $\mathbf{R}^{-\top} \hat{\mathbf{K}_{\mathbf{y}}}'' \mathbf{R}^{-1}$ has at most $n_c + 1$ distinct eigenvalues: either the eigenvalue is $1$, or it is equal to $1 + \lambda$ with $\lambda$ being some eigenvalue of \Cref{krep}.

Following the convergence analysis in \cite{shewchuk1994introduction}, if we assume infinite floating-point precision, the CG procedure with the system \Cref{precond kypp} converges in at most $n_c + 1$ iterations. This is a significant improvement over the original system $\hat{\mathbf{K}_{\mathbf{y}}}''$, which could contain up to $n_{\text{train}}$ distinct eigenvalues, and could take theoretically up to $n_{\text{train}}$ iterations to converge. We further demonstrate the fast convergence rate of PCG by experiments in \Cref{experiment}.

Combining all methods mentioned in this section, the pseudo-code presented in \Cref{pseudocode} shows the overall training process implemented in our framework.

\begin{algorithm}[h!]
\caption{Training loop of nuGPR} \label{pseudocode}
\begin{algorithmic}[1]
    \STATE{\textbf{procedure} \textsc{ComputeLoss}($\mathbf{R}$, $\lambda$, $\sigma^2$, $\alpha$, Equation $(e)$)}
        \STATE{\INDENT \textbf{if} Equation $(e)$ is provided}
            \STATE{\INDENT\INDENT adjust PCG left-multiplication according to Equation $(e)$}
        \STATE{\INDENT \textbf{end if}}
        \STATE{\INDENT solve $\hat{\mathbf{K}_{\mathbf{y}}}'' \mathbf{u} = \mathbf{y}_{\text{train}}$ for $\mathbf{u}$ by PCG with preconditioner $\mathbf{R}^{-1}$}
        \STATE{\INDENT compute $|\hat{\mathbf{K}_{\mathbf{y}}}''|$ according to \Cref{logdet} by PCG with preconditioner $\mathbf{R}^{-1}$}
        \STATE{\INDENT $\mathcal{L}\gets \mathbf{y}_{\text{train}}^\top \mathbf{u} + |\hat{\mathbf{K}_{\mathbf{y}}}''| + n_{\text{train}} \log(2\pi)$}
        \STATE{\INDENT \textbf{return} $\frac{\mathcal{L}}{2}$}
    \STATE{\textbf{end procedure}}
    \STATE{}
    \STATE{initialize GP model with $\theta = (\lambda, \sigma^2, \alpha)$}
    \STATE{initialize step sizes $\Delta\lambda$, $\Delta\sigma^2$, $\Delta\alpha$ as appropriate}
    \FOR{$i = 1$ to $50$}
        \STATE{compute $\mathbf{R}$ by block-diagonal Cholesky factorization on $\mathbf{K}_{\text{diag}}$}
        \STATE{$\mathcal{L}_0\gets$ \textsc{ComputeLoss}($\mathbf{R}$, $\lambda$, $\sigma^2$, $\alpha$, \Cref{linear1})}
        \STATE{$\frac{\partial\mathcal{L}}{\partial\lambda}\gets \infty$, $\frac{\partial\mathcal{L}}{\partial\sigma^2}\gets \infty$, $\frac{\partial\mathcal{L}}{\partial\alpha}\gets \infty$}
        \WHILE{$\frac{\partial\mathcal{L}}{\partial\lambda}$ does not meet numerical gradient threshold}
            \STATE{$\mathcal{L}_1\gets$ \textsc{ComputeLoss}($\mathbf{R}$, $\lambda + \Delta\lambda$, $\sigma^2$, $\alpha$)}
            \STATE{$\frac{\partial\mathcal{L}}{\partial\lambda}\gets \frac{\mathcal{L}_1 - \mathcal{L}_0}{\Delta\lambda}$, $\Delta\lambda \gets \frac{\Delta\lambda}{2}$}
        \ENDWHILE
        \WHILE{$\frac{\partial\mathcal{L}}{\partial\sigma^2}$ does not meet numerical gradient threshold}
            \STATE{$\mathcal{L}_1\gets$ \textsc{ComputeLoss}($\mathbf{R}$, $\lambda$, $\sigma^2 + \Delta\sigma^2$, $\alpha$, \Cref{linear2})}
            \STATE{$\frac{\partial\mathcal{L}}{\partial\sigma^2}\gets \frac{\mathcal{L}_1 - \mathcal{L}_0}{\Delta\sigma^2}$, $\Delta\sigma^2 \gets \frac{\Delta\sigma^2}{2}$}
        \ENDWHILE
        \WHILE{$\frac{\partial\mathcal{L}}{\partial\alpha}$ does not meet numerical gradient threshold}
            \STATE{$\mathcal{L}_1\gets$ \textsc{ComputeLoss}($\mathbf{R}$, $\lambda$, $\sigma^2$, $\alpha + \Delta\alpha$, \Cref{linear3})}
            \STATE{$\frac{\partial\mathcal{L}}{\partial\alpha}\gets \frac{\mathcal{L}_1 - \mathcal{L}_0}{\Delta\alpha}$, $\Delta\alpha \gets \frac{\Delta\alpha}{2}$}
        \ENDWHILE
        \STATE{update $\theta = (\lambda, \sigma^2, \alpha)$ by Adam optimizer based on the gradients}
    \ENDFOR
\end{algorithmic}
\end{algorithm}

\section{GPU Implementation} \label{gpu}
The nuGPR framework is designed to perform the entire GP inference on the GPU. It is written in C++ and depends solely on the CUDA Toolkit and a few other utility libraries for I/O and testing. We make the following design choices to ensure a highly efficient GPU implementation:

\begin{itemize}[left=0.25in]
    \item Other than loading the input data and fetching the prediction results, only $\mathcal{O}(1)$ memory transfer between CPU and GPU is performed in any intermediate steps. This ensures minimal latency and maximum throughput, and allows the GPU to always maintain high utilization.

    \item To best demonstrate the potential of GPU computation, we choose our input data to always be $16$-byte aligned, and we control the cluster size $b$ to be uniform for all clusters in each individual dataset. These choices ensure optimal memory alignment and cache behavior, minimize warp divergence, and facilitate load balancing for batch APIs, all of which are critical GPU-specific optimizations \cite{luebke2008cuda}. While minor modifications could be made to our code to support irregular input sizes, such changes would exacerbate the inherent limitations of GPUs, leading to a less fair and accurate portrait of nuGPR's capabilities.

    \item The majority of our linear algebra operations are mapped directly into APIs in cuBLAS and cuSOLVER, as summarized in \Cref{directapis}. For some operations that do not have explicit APIs, we translate them into one or more collective APIs to achieve an equivalent result. These operations are listed in \Cref{translateapis}.

    \item Finally, certain tasks, such as the generation of covariance matrices, cannot be effectively mapped into any library API(s). For these cases, we develop custom CUDA kernels, and ensure that our kernels operate near their maximum theoretical throughput, or Speed-of-Light (SoL) \cite{luebke2008cuda}. As an example, we consider our \texttt{\detokenize{generate_covar}} kernel to compute the covariance matrices for GPR an optimal implementation, since it achieves $85\%$ SoL in compute throughput, and $77\%$ SoL in memory throughput according to the profiling results reported by NVIDIA Nsight Compute\footnote{Software available on NVIDIA website: \url{https://developer.nvidia.com/nsight-compute}}.
\end{itemize}

\begin{table}[h!]
    \centering
    \begin{adjustbox}{width=\textwidth}
    \begin{tabular}{ccc}
        \hline
        Description & Mapped library API & Comments \\
        \hline\hline
        Scale a vector, $\mathbf{x}\gets a\mathbf{x}$ & \texttt{cublasSscal} & \\
        Add a scaled vector to another vector, $\mathbf{y}\gets \mathbf{y} + a\mathbf{x}$ & \texttt{cublasSaxpy} & \\
        Dot product of two vectors, $\mathbf{r}\gets \mathbf{x}\cdot \mathbf{y}$ & \texttt{cublasSdot} & \\
        Matrix add, $\mathbf{C}\gets a\mathbf{A} + b\mathbf{B}$ & \texttt{cublasSgeam} & To transpose $\mathbf{A}$: $\mathbf{C}\gets \mathbf{0} + \mathbf{A}^\top$ \\
        Matrix multiply and accumulate, $\mathbf{C}\gets a\mathbf{A}\mathbf{B} + b\mathbf{C}$ & \texttt{cublasSgemm} & \\
        Block-diagonal matrix multiply, $\mathbf{C}\gets \mathbf{K}_{\text{diag}}\mathbf{A}$ & \texttt{cublasSgemmStridedBatched} & Used in PCG for $\mathbf{K}_{\text{diag}}$ \\
        Cholesky factorization $\mathbf{A} = \mathbf{R}^\top \mathbf{R}$ & \texttt{cusolverDnXpotrf} & Called on diagonal blocks $(\mathbf{K}_{\text{diag}})_i$ \\
        \hline
    \end{tabular}
    \end{adjustbox}
    \caption{Directly mapped APIs for linear algebra operations}
    \label{directapis}
\end{table}

\begin{table}[h!]
    \centering
    \begin{adjustbox}{width=\textwidth}
    \begin{tabular}{ccc}
        \hline
        Description & Translated library API(s) & Comments \\
        \hline\hline
        Scale column vectors in $\mathbf{C}$ by coefficients in $\mathbf{x}$ & \texttt{cublasSdgmm} & $\mathbf{C}\gets \mathbf{A}\cdot \text{diag}(\mathbf{x})$ \\
        Batch dot product of all $m$ columns in $\mathbf{A}$ and $\mathbf{B}$ & \texttt{cublasSdgmm}, then \texttt{cublasSgemm} & Flatten both $\mathbf{A}$ and $\mathbf{B}$, then \\
        & & $\mathbf{r}\gets \mathbf{1}^\top\cdot \text{flatten}(A)\cdot \text{diag}(\text{flatten}(B))$ \\
        Block triangular solve $\mathbf{C}\gets \mathbf{R}^{-1} \mathbf{A}$ & \texttt{cusolverDnXtrtri}, then & Obtain $\mathbf{R}_i^{-1}$ by triangular inverse, \\
        & \texttt{cublasSgemmStridedBatched} & then use batched multiply \\
        \hline
    \end{tabular}
    \end{adjustbox}
    \caption{Translated APIs for linear algebra operations}
    \label{translateapis}
\end{table}

To summarize our strategy, we adhere to the principle of maximizing GPU utilization. By keeping the GPU close to $100\%$ utilized, we fully exploit its parallel processing capabilities. In practice, we use NVIDIA Nsight Systems\footnote{Software available on NVIDIA website: \url{https://developer.nvidia.com/nsight-systems}} to monitor GPU utilization during critical procedures to validate our approach. For example, Nsight Systems can be used to benchmark one iteration of our batched-CG procedure, where we observe that except for small gaps between kernel invocations, the GPU utilization rate is almost always close to $100\%$.

We also highlight a few distinct advantages of our C++/CUDA based implementation compared to existing frameworks using TensorFlow or PyTorch. First, the block-diagonal structure of matrices such as $\mathbf{K}_{\text{diag}}$ requires us to use the strided-batched GEMM API from cuBLAS, which offers much faster performance over computing individual GEMMs sequentially. Also, to obtain $\mathbf{R}$, we compute $n_c$ Cholesky factorizations on $b\times b$ matrices in a batch by leveraging CUDA streams in cuSOLVER API calls. Above choices are made with both \Cref{pseudocode}'s characteristics and CUDA's language features in mind, and are most easily realized by interfacing with CUDA directly in C++, rather than using any intermediate framework.

\section{Experimental Results} \label{experiment}
\subsection{Datasets}
We use a variety of synthetic and real-world datasets to benchmark the performance of our nuGPR framework. Because we need to supply cluster information along with input data, one key challenge is to organize our datasets to expose such properties clearly. This is addressed by either generating synthetic data based on predefined cluster patterns, or preprocessing real-world data based on techniques and criteria known to help with clustering.

\subsubsection*{Synthetic Datasets}
The simplest datasets to test our implementation are one-dimensional in the form $\mathbf{y}_i = f(\mathbf{x}_i)$ where $\mathbf{x}_i$ is a real number and $f: \mathbb{R} \to \mathbb{R}$. Once $f$ is specified, we choose $n_c$ equally spaced representative points on the $\mathbf{x}$-axis, and generate $b$ points randomly within a selected radius from each $\mathbf{r}_i$ to form non-overlapping clusters for $\mathbf{X}_{\text{train}}$. Finally, we select two values $m_1 < \min(\mathbf{X}_{\text{train}})$ and $m_2 > \max(\mathbf{X}_{\text{train}})$, and generate $\mathbf{X}_{\text{test}}$ uniformly between $(m_1, m_2)$, since uniformly distributed test points provide more meaningful plots of GP prediction means and variances.

The same idea can be extended to a multi-dimensional dataset, where now $\mathbf{x}_i = (x_1, \ldots, x_d)$ is a vector with $d > 1$, and $f: \mathbb{R}^d \to \mathbb{R}$. To construct the array $\mathbf{r}$, we divide the $d$-dimensional space into hypercubes with a fixed side length $l$, and require $\mathbf{r}_i$ to be a vertex of a hypercube. For $\mathbf{X}_{\text{train}}$, any cluster radius less than $\frac{l}{2}$ can ensure the input clusters do not overlap. For $\mathbf{X}_{\text{test}}$, since we do not intend to plot the dataset, picking random points based on the data in $\mathbf{X}_{\text{train}}$ should suffice for verification purposes.

For simplicity, we pick one sample function in each dimensional category to generate our synthetic datasets. The one-dimensional dataset uses the function
\begin{equation} \label{eq1d}
    f(\mathbf{x}_i) = \frac{\sin(2\mathbf{x}_i)}{\mathbf{x}_i} + \sigma_i^2,\ \sigma_i^2\sim \mathcal{N}(0, 0.16),
\end{equation}
and the three-dimensional dataset uses
\begin{equation} \label{eq3d}
    f(\mathbf{x}_i) = \frac{x_1^2 + x_2^2 + x_3^2}{100} + \sigma_i^2,\ \sigma_i^2\sim \mathcal{N}(0, 0.16),
\end{equation}
where $\mathbf{x}_i = (x_1, x_2, x_3)$, and $\sigma_i^2$ is an additional noise term sampled from a normal distribution.

\subsubsection*{Real-World Datasets}
Three real-world datasets: Kin40k \cite{nguyen2014fast}\footnote{The dataset is obtained from the code repository of the cited paper: \url{https://github.com/trungngv/fgp}}, Gas Sensor \cite{misc_gas_sensor_array_drift_at_different_concentrations_270}, and MNIST \cite{deng2012mnist} are used to additionally analyze the performance of nuGPR.

\textbf{The Kin40k dataset} consists of $40\text{,}000$ samples, and each sample includes $8$ input features and $1$ output label describing kinematic operations. We first use Principle Component Analysis (PCA) on the normalized input to extract $d = 4$ components per sample, and then apply k-means clustering to find the $n_c = 10$ clusters we need for GP inference. Since k-means does not guarantee even cluster distribution, certain data points need to be dropped for us to pick a uniform cluster size $b$. We observe $b = 1\text{,}600$ to be a feasible cluster size, and hence the eventual $\mathbf{X}_{\text{train}}$ has $n_{\text{train}} = 16\text{,}000$. The construction of $\mathbf{r}$ is straightforward, as we can just use the cluster centers reported by k-means. Finally, we randomly select $n_{\text{test}} = 4\text{,}000$ samples from the rest of the dataset to form $\mathbf{X}_{\text{test}}$.

\textbf{The Gas Sensor dataset} consists of $13\text{,}910$ samples, and each sample contains $128$ features collected from $16$ sensors, along with information about the underlying gas type ($6$ in total), its concentration level, and time of measurement. We formulate this as a regression task, where the concentration level is the output. Similar to Kin40k, we first use PCA on the normalized input, except here we extract $16$ components to align with the number of sensors. The gas type and measurement time are also normalized, and are concatenated as another $2$ columns in the input data -- making the total dimension $d = 18$.

The clustering of the gas sensor data can be addressed by grouping the samples based on gas type, so we have $n_c = 6$ clusters. To accommodate for the gas type imbalance in the original dataset and the need to have a substantial sized $\mathbf{X}_{\text{test}}$, we pick $b = 1\text{,}600$ to form $\mathbf{X}_{\text{train}}$ with $n_{\text{train}} = 9\text{,}600$. There is no direct information for constructing $\mathbf{r}$, so we set each
\begin{equation}
    \mathbf{r}_i = \mathop{\mathrm{argmin}}_{\mathbf{x}} \sum_{x'\in \mathbf{X}_i} k(\mathbf{x}, \mathbf{x}'),\ i = 1,\ldots,n_c,
\end{equation}
following the intuition that the best representative point in a cluster has the smallest total kernel distance from all other points in the same cluster. Finally, we collect $n_{\text{test}} = 2\text{,}400$ samples randomly from the remaining pool to form $\mathbf{X}_{\text{test}}$.

\textbf{The MNIST dataset} consists of $60\text{,}000$ training samples and $10\text{,}000$ testing samples, where each sample is an $28\times 28$ grayscale image of a handwritten digit $0$-$9$. We employ a similar technique as \cite{wilson2016deep} to fine-tune a ResNet-18 \cite{he2016deep} and extract the required features for GP from images. Specifically, we replace the last $512\times 1000$ linear layer with 3 layers: a $512\times 10$ linear layer to extract $d = 10$ features for GP, a ReLU layer to add non-linearity, and a $10\times 10$ linear layer to map to the output classes. We only train the additional layers for $50$ epochs on all training samples.

MNIST is inherently clustered by digit classes; however, if we cluster by each digit, the prediction task for GPR becomes trivial since the ground truth is the same as the cluster ID. To devise another form of clustering, we randomly augment the input images with rotations in increments of $45$ degrees and horizontal flips to create a total of $n_c = 8\times 2 = 16$ clusters. Here, for $\mathbf{X}_{\text{train}}$, we select $n_{\text{train}} = 20\text{,}000$ and $b = 1\text{,}250$. Note that a larger $n_{\text{train}}$ could have been chosen for nuGPR, however comparison with GPyTorch would have been impossible since its training procedure would have required memory that exceeds our GPU's physical limit. Finally, $\mathbf{X}_{\text{test}}$ contains $n_{\text{test}} = 5\text{,}000$ samples from all testing samples, maintaining the same train-test split as above.

\begin{table}[h!]
    \centering
    \begin{adjustbox}{width=\textwidth}
    \begin{tabular}{cccccccc}
        \hline
        Dataset name & Samples & Processing method & Output & $n_{\text{train}}$ & ${n_{\text{test}}}$ & $d$ & $n_c$ \\
        \hline\hline
        Synthetic 1-D & N/A & Generate around chosen centers & Result of $f$ & $4\text{,}000$ to $20\text{,}000$ & $1\text{,}000$ to $5\text{,}000$ & $1$ & $10$ \\
        Synthetic 3-D & N/A & Generate around hypercube vertices & Result of $f$ & $4\text{,}000$ to $20\text{,}000$ & $1\text{,}000$ to $5\text{,}000$ & $3$ & $10$ \\
        Kin40k & $40\text{,}000$ & PCA + k-means & Kinematics label & $16\text{,}000$ & $4\text{,}000$ & $4$ & $10$ \\
        Gas Sensor & $13\text{,}910$ & PCA + group by gas type & Gas concentration & $9\text{,}600$ & $2\text{,}400$ & $18$ & $6$ \\
        MNIST & $70\text{,}000$ & ResNet-18 and image augmentation & Digit class $0$-$9$ & $20\text{,}000$ & $5\text{,}000$ & $10$ & $16$ \\
        \hline
    \end{tabular}
    \end{adjustbox}
    \caption{Summary of datasets used for benchmarking nuGPR}
    \label{dataset}
\end{table}

\Cref{dataset} gives a summary of the specifications for all synthetic and real-world datasets we've discussed above.

\subsection{Experimental Setup}
\subsubsection*{Baseline}

Both GPFlow and GPyTorch mentioned in \Cref{related work} support GP inference on the GPU, but according to \cite{gardner2018gpytorch}, the CG-based exact GP implementation of GPyTorch outperforms the Cholesky-based implementation from GPFlow across all datasets, in terms of both performance and test accuracy. Therefore, we select GPyTorch as our baseline for testing nuGPR, and compare the computation speed, memory consumption, and RMSE for both frameworks under the selected datasets.

\subsubsection*{Premise}
We process our datasets into a group of \texttt{.npy} files read by both nuGPR and GPyTorch, to ensure both frameworks consume the exact same input data. Both models are trained for $50$ epochs, and two metrics are measured: the total time spent in training, as well as the memory usage. We choose the learning rate as $\gamma = 0.05$ for the Adam optimizer. The initial hyperparameters are randomly selected before each training process, but both models always start from the exact same initial values. For test accuracy, we implement for nuGPR an equivalent posterior forward routine as GPyTorch, and we calculate the RMSE against the output labels for both frameworks under the same test data.

Regarding model-specific settings, we choose to terminate our PCG procedure when the L2 norm of the residual vector is less than $0.01$. We intentionally choose a large value $2\text{,}000$ as the maximum number of iterations for CG, to flag any instances when CG fails to converge by our threshold. We use $m = 8$ Hutchinson vectors in the estimation of $\log|\hat{\mathbf{K}_{\mathbf{y}}}|$ for nuGPR, since $m = 8$ satisfies the memory alignment requirement for GPU computation, while any larger value of $m$ does not yield a noticeably better result. Such vectors are also pre-generated as one of the \texttt{.npy} files. For GPyTorch, we maintain all default settings\footnote{The default settings can be found under the source code for GPyTorch: \url{https://github.com/cornellius-gp/gpytorch/blob/master/gpytorch/settings.py}} to portray the framework’s out-of-the-box performance, ensuring the most fair comparison against our work.

\subsubsection*{Test Platform}
All test results are collected on a Ubuntu 22.04 desktop with an AMD R9-5950x 16-core CPU, 32 GB RAM, and an NVIDIA RTX 3080 Ti GPU with 12GB VRAM. For the nuGPR framework, we compile the project with CMake 3.28.3, GCC 11.4 and CUDA Toolkit 11.8, and we supply the compiler flags \texttt{-O3} and \texttt{-std=gnu++17}. For our baseline, we use the newest v1.11 release of GPyTorch in a Conda environment with Python 3.11 and PyTorch 2.1.2 (CUDA support enabled). All computation is performed in single-precision.

The timing information is collected by \texttt{\detokenize{std::chrono::high_resolution_clock}} for nuGPR, and by Python's built-in \texttt{time} module for GPyTorch. The memory consumption is measured by monitoring the counters provided by \texttt{\detokenize{nvidia-smi}}. While more advanced tools exist for profiling native CUDA programs \cite{luebke2008cuda}, these tools are often less compatible with frameworks like PyTorch, which indirectly calls into CUDA APIs. Therefore, our proposed measurement techniques offer a more fair and straightforward method for comparing the performance of the two frameworks.

\subsection{Results and Analysis}
\begin{figure}[h!]
    \centering
    \includegraphics[width=\textwidth]{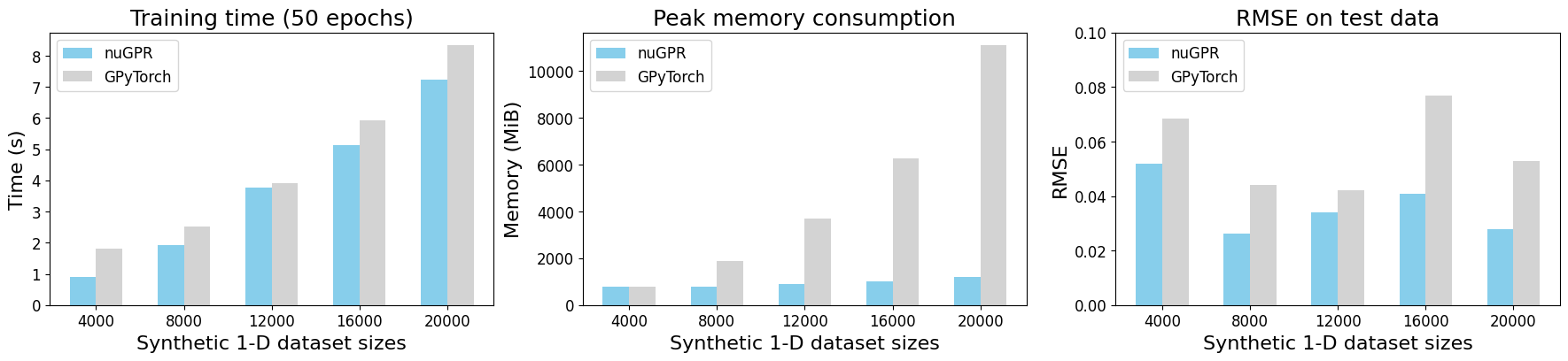}
    \caption{Results of synthetic 1-D dataset defined in \Cref{eq1d}}
    \label{fig1d}
\end{figure}

\Cref{fig1d} compares the total training time, peak memory consumption, and RMSE score for the two frameworks using the synthetic 1-D dataset defined by \Cref{eq1d}. Inputs with $n_{\text{train}} = 4\text{,}000$ to $n_{\text{train}} = 20\text{,}000$ are generated in increments of $4\text{,}000$, until our baseline GPyTorch runs out of VRAM to complete the GP inference. Subsequently, $n_{\text{test}}$ is chosen to maintain a train-test split at $80\%$ training to $20\%$ testing, which is consistent for all our datasets.

We can observe from \Cref{fig1d} that the advantage in training time is decisive for nuGPR on smaller input sizes, but the gap shrinks when $n_{\text{train}}\geq 12\text{,}000$. Nevertheless, nuGPR is able to maintain roughly the same rate-of-growth compared to GPyTorch, meaning the two algorithms scale equally well against increasing problem sizes. For memory consumption, nuGPR not only outperforms GPyTorch on every input size, but also maintains a lower rate-of-growth, hence the memory performance of nuGPR scales better than GPyTorch. Finally, the RMSE scores for nuGPR are consistently lower than those for GPyTorch, as expected, given that nuGPR is designed to be cluster-aware, whereas GPyTorch assumes more generic input data. The perfectly exposed cluster patterns in our synthetic data naturally benefit our framework to a greater extent than GPyTorch.

\begin{figure}[h!]
    \centering
    \includegraphics[width=\textwidth]{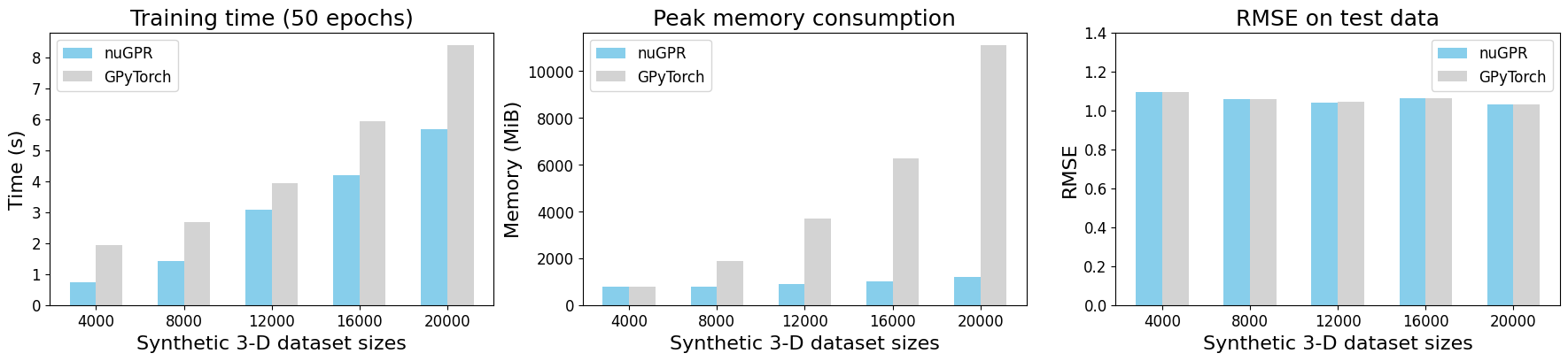}
    \caption{Results of synthetic 3-D dataset defined in \Cref{eq3d}}
    \label{fig3d}
\end{figure}

In \Cref{fig3d}, we capture the same metrics for the synthetic 3-D dataset defined by \Cref{eq3d}. The choices for $n_{\text{train}}$ and $n_{\text{test}}$ remain the same as 1-D. Here, the conclusions for training time and memory consumption stay consistent, and we point out that the memory consumption data is almost identical to 1-D case, since the dominating factor of storage in GP inference is the covariance matrix, whose size is only affected by $n_{\text{train}}$. The RMSE scores are almost tied between nuGPR and GPyTorch, which means the two frameworks exploit cluster patterns about equally well in the 3-D case.

\begin{figure}[h!]
    \centering
    \includegraphics[width=\textwidth]{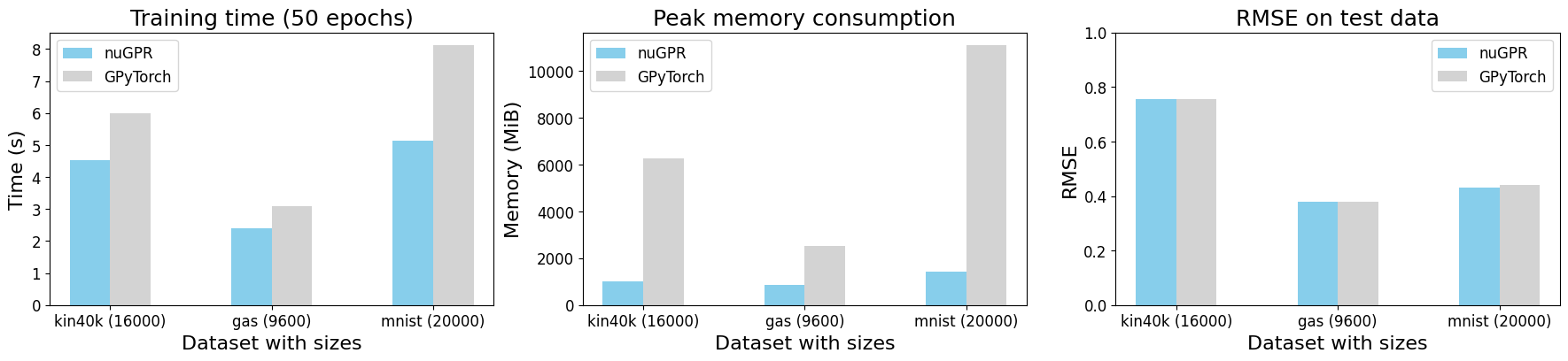}
    \caption{Results of real-world datasets (Kin40k, Gas Sensors, MNIST)}
    \label{figreal}
\end{figure}

The metrics for all $3$ real-world datasets are displayed in \Cref{figreal}. Our preprocessing steps facilitate clustering across various datasets, and as a result, nuGPR demonstrates a performance lead over GPyTorch on real-world data as much as on synthetic data. In terms of RMSE scores, the two frameworks are still closely matched. An important observation is that despite the collection of real-world data being less controlled than the generation of synthetic data, the test errors of the two frameworks consistently remain within a small margin of each other.

\begin{figure}[h!]
    \centering
    \includegraphics[width=\textwidth]{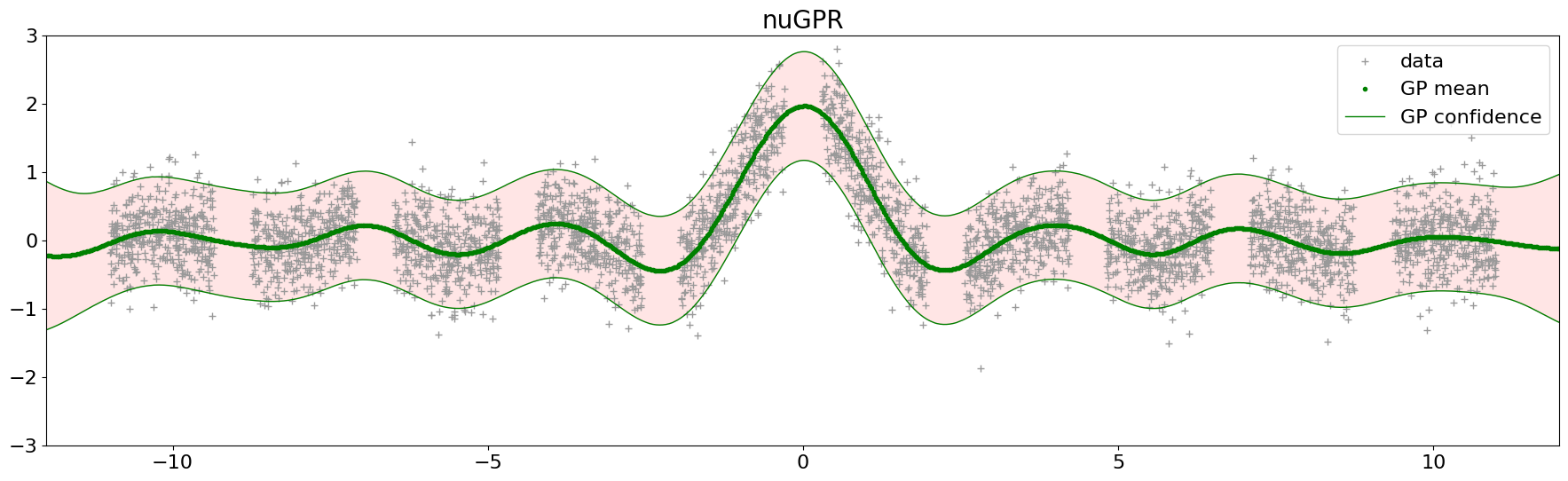}
    \includegraphics[width=\textwidth]{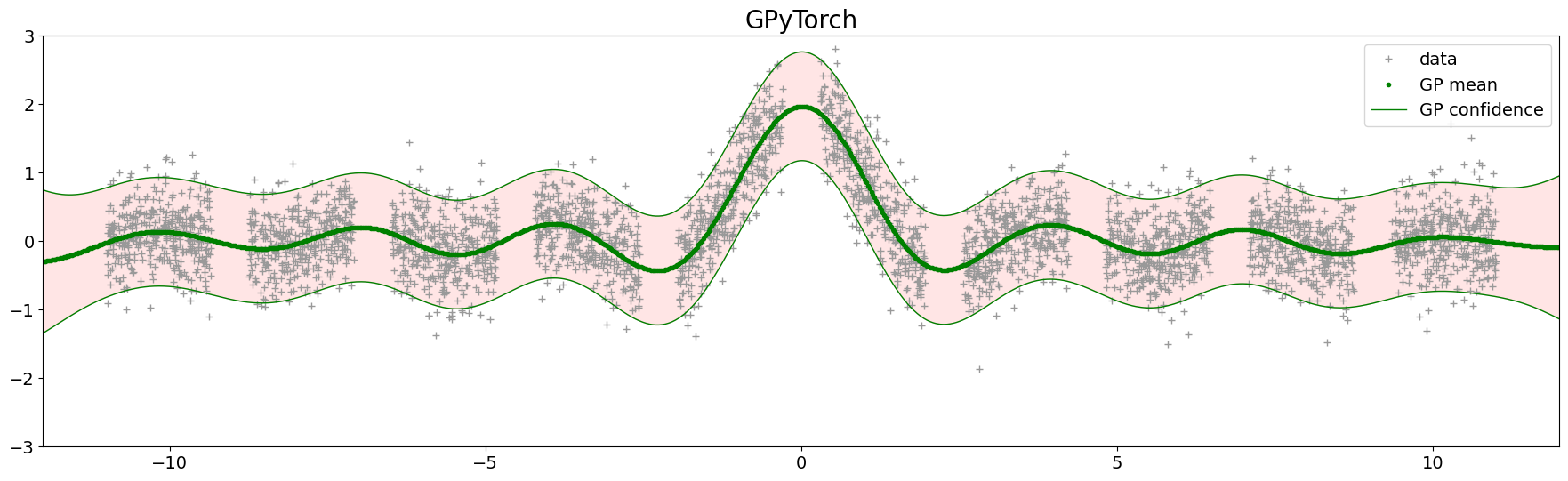}
    \caption{GP predictions for synthetic 1-D dataset. The green bold lines denote the mean, and the pink shaded areas denote the confidence region. The nuGPR and GPyTorch frameworks result in nearly identical predictions.}
    \label{figpred}
\end{figure}

\begin{figure}[h!]
    \centering
    \includegraphics[width=\textwidth]{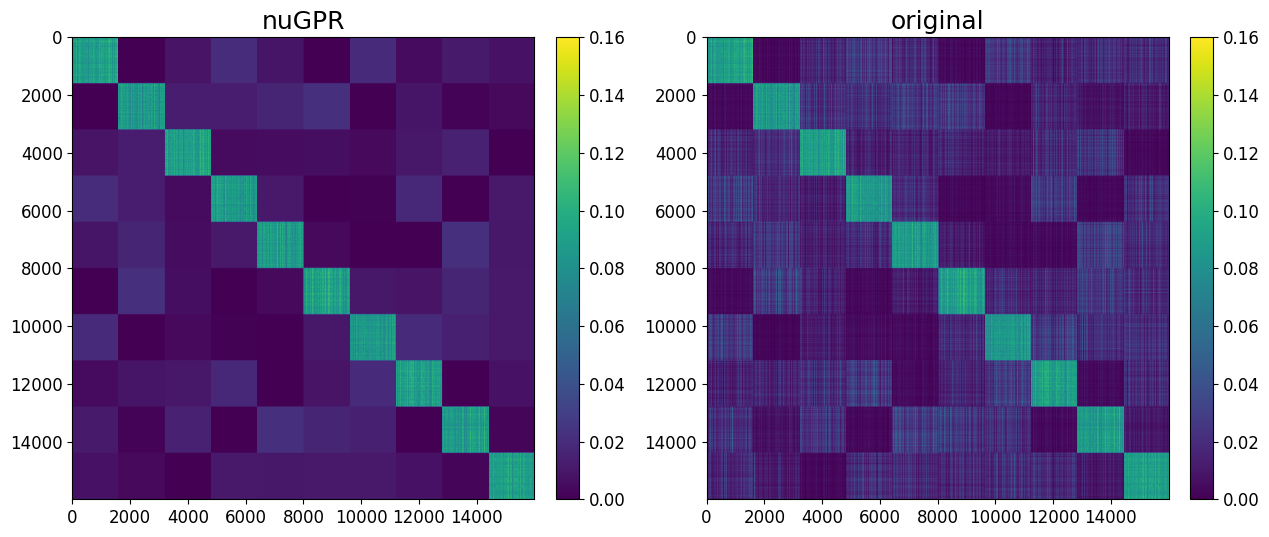}
    \vspace{-0.1in}
    \caption{Covariance matrices after 50 epochs for Kin40k dataset. The matrix from nuGPR contains unchanged diagonal blocks and approximated off-diagonal blocks. The overall pattern in the original matrix is preserved well.}
    \label{figcovar}
\end{figure}

We also present visualization of the two GPR frameworks following some commonly used techniques to interpret the predictions \cite{he2016deep, saatcci2012scalable, wang2023intuitive}. \Cref{figpred} plots the prediction mean and confidence region after training on the 1-D dataset with $n_{\text{train}} = 4\text{,}000$. Both frameworks achieve similar regression results without any prior knowledge on $f$. \Cref{figcovar} plots the covariance matrices as heatmaps from both frameworks after training, demonstrating how nuGPR keeps the diagonal blocks unchanged, while replacing each off-diagonal block with a single entry (rank-$1$ approximation), as detailed in \Cref{low rank}.

\begin{figure}[h!]
    \centering
    \includegraphics[width=\textwidth]{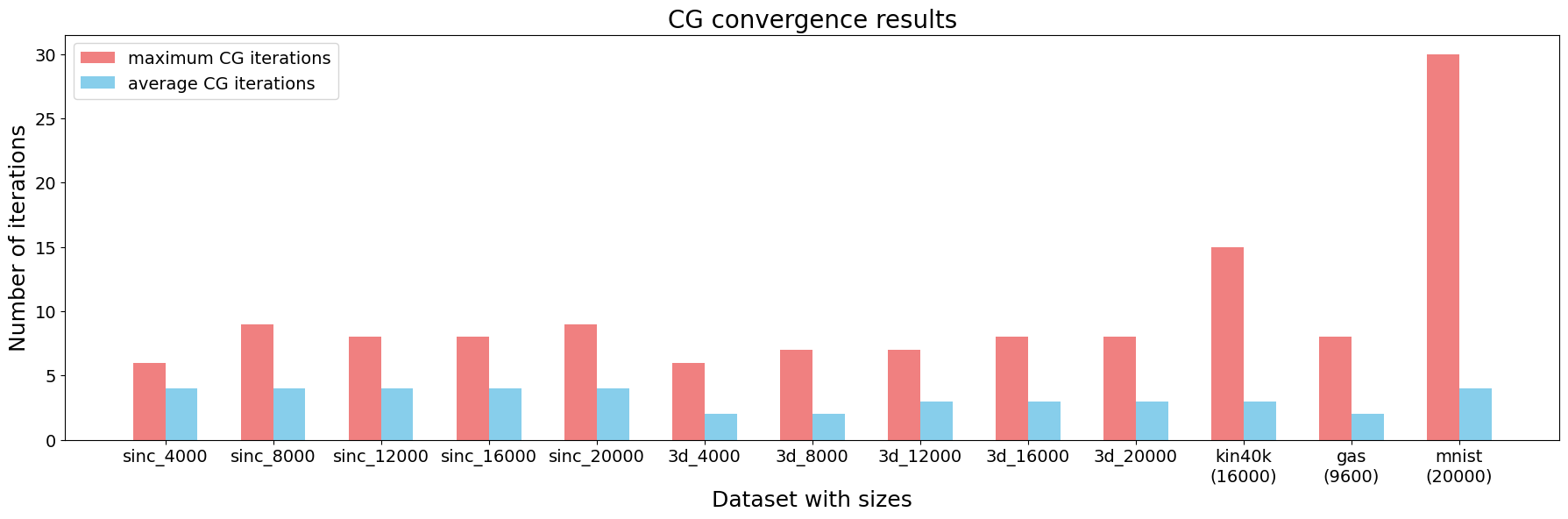}
    \vspace{-0.1in}
    \caption{Maximum and average number of CG iterations for nuGPR}
    \label{figcg}
\end{figure}

Lastly, to study the impact of preconditioning, we log the number of PCG iterations performed in nuGPR on each dataset across $50$ epochs of training. \Cref{figcg} presents the maximum and average number of PCG iterations for all datasets. The results align well with our theoretical upper bound of $n_c + 1$ iterations established in \Cref{low rank}; for certain datasets, more iterations than $n_c + 1$ are required, which is due to our usage of only single-precision floating point numbers.

\section{Discussion} \label{discussion}
We highlight the main advantages of nuGPR as a novel Gaussian Process Regression framework:

\begin{itemize}[left=0.25in]
    \item \textbf{Effectiveness of preconditioning:} We apply the preconditioner $\mathbf{R}^{-1}$ based on the assumption that the diagonal blocks capture the most important information within $\hat{\mathbf{K}_{\mathbf{y}}}$, and we show that it significantly helps accelerate the convergence of PCG especially when the data is well-clustered. As illustrated in \Cref{figcg}, the average number of CG iterations is less than $5$ for all our datasets. This includes the solves required for \Cref{PQ}, which can converge more slowly due to the increased condition number from $Q(\hat{\mathbf{K}_{\mathbf{y}}})$. Despite this difficulty, the adaptation of $\mathbf{R}^{-1}$ still helps maintain a low average number of iterations. Even the global maximum of $30$ iterations does not substantially impact our performance. It is evident that the rapid computation speed of nuGPR is closely linked to the above preconditioning strategy.

    \item \textbf{Performance benefit of low-rank approximations:} As demonstrated both theoretically and through experiments, the adoption of low-rank approximation $\hat{\mathbf{K}_{\mathbf{y}}}'$ improves both speed and memory consumption for nuGPR. Furthermore, the RMSE data shows this performance increase is not a result of sacrificing test accuracy in both synthetic and real-world data. Such advantages, especially the reduction in memory usage by over $12\times$, enable single-GPU training of GPR on very large datasets -- a task not possible with existing GP libraries.

    \item \textbf{A CG-only approach for GPR training:} By employing the methods in \Cref{methodology} and eliminating back-propagation through numerical gradient, we have developed a GPR training routine whose results depend only on the conjugate gradient method. Although the computation of \Cref{cholesky factorization} relies on Cholesky factorization, we note that preconditioning does not alter the output of CG beyond any floating point errors. Consequently, nuGPR is able to achieve both high computational efficiency and accuracy, and we attribute this success to the numerous favorable properties of CG, such as being numerically stable, parallelization-friendly, and straightforward to implement \cite{shewchuk1994introduction}.

    \item \textbf{Optimal GPU computation with minimal dependency:} We achieve great performance for large-scale GPR training on GPUs by leveraging the best linear algebra libraries and applying various optimizations. Relying solely on the CUDA Toolkit, nuGPR minimizes its dependencies and avoids the overheads associated with other frameworks \cite{GPflow2017, gardner2018gpytorch}, such as JIT compilation. This streamlined approach helps nuGPR exploit full capabilities of NVIDIA hardware, while remaining highly portable as a software framework.
\end{itemize}

We recognize that the clustering properties of input data are a key assumption in the nuGPR framework. A significant portion of nuGPR's performance gains relies on the distribution of input samples into uniform-sized clusters, which enables us to replace off-diagonal blocks (representing correlation matrices between samples in two clusters) with low-rank approximations. In our opinion, clustering/blocking provides a tunable, flexible mechanism to reduce the memory requirement and improve computational throughput of GPR, especially when the datasets are favorably clustered. In the absence of inherent clustering in a dataset, one can apply manually introduced clustering approaches to extract such benefits to some extent, as demonstrated by our experiments with real-world datasets such as Kin40k. As a result, our method remains broadly applicable to most datasets in the studies of GPR.

Lastly, we also identify several future directions for enhancing the nuGPR framework. First, alternative preconditioning strategies that do not rely on Cholesky factorization, such as using Singular Value Decomposition, could be explored to provide additional performance improvements. And second, while nuGPR can handle very large datasets on a single GPU due to its memory efficiency, extending it to multi-GPU training scenarios would push its performance boundary even further.

\section{Conclusion} \label{conclusion}
In this paper, we introduce a new framework for Gaussian Process Regression training based on iterative algorithms and off-diagonal low-rank approximations. Our framework is designed to operate entirely on GPUs, capitalizing on cutting-edge hardware capabilities as well as the CUDA Toolkit and its relevant libraries. The underlying GPR model is optimized through numerical gradient to avoid the computation needed for differentiating the hyperparameters. We use the Pad\'e approximation to convert log determinant computation into a linear solve, and we describe a preconditioning strategy based on diagonal blocks of the covariance matrix, which greatly speeds up convergence of the conjugate gradient method. By selecting representative points based on clusters and using low-rank approximations, we notably reduce our training time and memory usage without compromising the model accuracy, which is demonstrated through our experiments on various synthetic and real-world datasets.

\bibliographystyle{siamplain}
\bibliography{references}
\end{document}

%% file: ex_shared.tex
\usepackage{lipsum}
\usepackage{amsfonts}
\usepackage{graphicx}
\usepackage{epstopdf}
\usepackage{algorithmic}
\ifpdf
  \DeclareGraphicsExtensions{.eps,.pdf,.png,.jpg}
\else
  \DeclareGraphicsExtensions{.eps}
\fi

\usepackage{enumitem}
\usepackage{algorithmic}
\usepackage{adjustbox}

\Crefname{ALC@unique}{Line}{Lines}

\newcommand{\INDENT}[1][1]{\hspace{#1\algorithmicindent}}

\newsiamremark{remark}{Remark}
\newsiamremark{hypothesis}{Hypothesis}
\crefname{hypothesis}{Hypothesis}{Hypotheses}
\newsiamthm{claim}{Claim}

\headers{nuGPR: GPU-Accelerated Gaussian Process Regression}
{Z. Zhao and V. Sarin}

\title{nuGPR: GPU-Accelerated Gaussian Process Regression with Iterative Algorithms and Low-Rank Approximations
\thanks{Submitted to the editors July 10, 2025.}}

\author{Ziqi Zhao\thanks{Department of Computer Science and Engineering, Texas A\&M University, College Station, TX 
  (\email{astrajoan@tamu.edu}, \email{sarin@tamu.edu}).}
\and Vivek Sarin\footnotemark[2]}

\usepackage{amsopn}